# Generative Fuzzy System for Sequence Generation

Hailong Yang, Zhaohong Deng, *Senior Member, IEEE*, Wei Zhang, Zhuangzhuang Zhao, Guanjin Wang, Kup-sze Choi, *Senior Member, IEEE*

*Abstract*—Generative Models (GMs), particularly Large Language Models (LLMs), have garnered significant attention in machine learning and artificial intelligence for their ability to generate new data by learning the statistical properties of training data and creating data that resemble the original. This capability offers a wide range of applications across various domains. However, the complex structures and numerous model parameters of GMs make the input-output processes opaque, complicating the understanding and control of outputs. Moreover, the purely data-driven learning mechanism limits GM's ability to acquire broader knowledge. There remains substantial potential for enhancing the robustness and generalization capabilities of GMs. In this work, we introduce the fuzzy system, a classical modeling method that combines data and knowledge-driven mechanisms, to generative tasks. We propose a novel Generative Fuzzy System framework, named GenFS, which integrates the deep learning capabilities of GM with the interpretability and dual-driven mechanisms of fuzzy systems. Specifically, we propose an end-to-end GenFS-based model for sequence generation, called FuzzyS2S. A series of experimental studies were conducted on 12 datasets, covering three distinct categories of generative tasks: machine translation, code generation, and summary generation. The results demonstrate that FuzzyS2S outperforms the Transformer in terms of accuracy and fluency. Furthermore, it exhibits better performance on some datasets compared to state-of-the-art models T5 and CodeT5.

## I. INTRODUCTION

In recent years, generative models have garnered widespread attention for addressing complex generative tasks. Particularly, the continuous development of Transformer and its derivative technologies has led to state-of-the-art large natural language processing models (LLMs), such as ChatGPT and Llama [1], [2], [3]. These models can explore hidden patterns and relationships within the data and generate high-quality multimodal data such as text, audio, and image, making them powerful tools in Natural Language Processing (NLP) and artificial intelligence.

Transformers [4] and their derivatives form the basis for many generative AI techniques. Representative models in this area include GPT [5], [6], [7], [8], [9] for text generation, CLIP [10], [11] for image generation, and MuLan [12], [13], [14] for audio generation. Despite the complexity and variety of generative tasks, they can be transformed into sequence generative tasks through serialization techniques. For example, text is sliced into token sequence, images are divided into patches, and audio data is discretized into time series of amplitudes. The models generate predicted sequences by learning the relationships between input and target sequences. These sequences are then deserialized in order to obtain the final results. Sequence generative tasks are characterized by several key attributes: they involve variable-length, unstructured data; the elements of a sequence are ordered; and the mapping relationship between input and target sequences is complex.

Existing generative models have complex network structures, deep hierarchies, and a large number of parameters, making their internal workings and decision-making processes opaque and their output difficult to control [15]. Also, these models often require vast datasets for training and operate as black boxes, making it hard to incorporate logical knowledge, rules and constraints. This data-driven approach limits the models' ability to generalize and handle a wider range of applications. Generative models also demand significant computational resources and time for training. Moreover, the trained models' parameters are challenging to reuse in other models, leading to a considerable waste of computational resources and time, which hinders their large-scale application and technological development from the long run.

The fuzzy system is a classical modeling method composed of fuzzy sets, fuzzy rules, and inference mechanisms. It effectively handles fuzzy information and provides good interpretability [16]. Also, fuzzy systems can express expert knowledge, address nonlinear problems, and offer better robustness and generalization capabilities. Notably, fuzzy systems capture the fuzzy characteristics of human brain thinking from a macroscopic perspective, simulating human reasoning and decision-making to handle uncertainty problems that conventional mathematical methods struggle to solve. Currently, fuzzy systems are widely used in tasks such as classification [17], [18], recognition [19], [20], and detection [21]. In fuzzy systems, fuzzy rules can represent priori expert knowledge, which can be easily transferred from one system to another [22], [23]. This enables efficient knowledge migration and knowledge reuse, greatly saving computational resources and training time.

The input sequences of generative tasks may exhibit widely varying characteristics, such as different lengths, different

This work was supported in part by the National Key R&D Program of China under Grant 2022YFE0112400, and in part by the National Natural Science Foundation of China under Grant 62176105. (Corresponding author: Zhaohong Deng).

H. Yang, W. Zhang, Z. Zhao and Z. Deng are with the School of Artificial Intelligence and Computer Science, and Engineering Research Center of Intelligent Technology for Healthcare, Ministry of Education, Wuxi 214122, China. (e-mail: yanghailong@stu.jiangnan.edu.cn; 7201607004@stu.jiangnan.edu.cn; zhaozhuangzhuang123@outlook.com; dengzhaohong@jiangnan.edu.cn;).

G. Wang is with the School of Information Technology, Murdoch University, WA, Australia (e-mail: Guanjin.Wang@murdoch.edu.au).

K. S. Choi is with the TechCosmos Ltd., Hong Kong. (e-mail: kschoi@ieee.org).

token frequency distributions, and inherent ambiguity and uncertainty. Fuzzy system can use a priori expert knowledge to divide the inputs into multiple sets with different feature terms for respective processing. Therefore, it is significant to develop a novel generative fuzzy system. However, for complex generative tasks, such as machine translation, code generation, and summary generation in NLP community, the generative fuzzy system is still difficult to cope with the following challenges: Firstly, the input of generative tasks consists of variable-length unstructured data, which cannot be directly processed by classical fuzzy systems. Secondly, the complex token mapping relationships require models to have a robust ability to handle high-dimensional data. However, classical fuzzy systems are shallow models with a small number of trainable parameters, making them inadequate for effectively undertaking generative tasks.

To address the first challenge, in the generative fuzzy system, we will implement delegate election for fuzzy sets and transform the fuzzy membership calculation into a similarity calculation between the inputs and the delegates. This approach enables the fuzzification of both structured and unstructured data. Generative tasks aim to model joint probability by learning the probability distribution of tokens across sequences, necessitating models with substantial learning capabilities and a large number of trainable parameters. To address the second challenge, we will introduce deep generative models as the consequents of the fuzzy rules. This integration enhances the system's learning ability, making it more suitable for sequence generation.

The generative fuzzy system retains the advantages of classical fuzzy systems, as both are fuzzy-rule-based systems known for their good interpretability and a dual-driven mechanisms powered by both knowledge and data. The main contributions of this paper can be summarized as follows:

Firstly, we propose a novel generative fuzzy system framework, namely GenFS. This framework combines the high interpretability of fuzzy system with the powerful learning capabilities of generative models. Moreover, GenFS can efficiently transfer generative knowledge from one system to another through generative fuzzy rules, thereby saving computational resources and reducing training time costs.

Secondly, based on our proposed GenFS framework, we introduce a specific generative fuzzy system for natural language sequence generation, called FuzzyS2S. During the preprocessing stage, FuzzyS2S employs a novel multi-scale fuzzy tokenizer to optimize the token frequency distribution of the sequences and to extract sequence information at multiple scales. In addition, the model introduces an innovative fuzzy membership calculation method, which effectively addresses the fuzzification problem of variable-length sequences.

The remainder of the paper is structured as follows: Part II introduces the generative model along with the related concepts and principles of classical fuzzy systems. Part III presents the generative fuzzy system framework, namely GenFS. Part IV presents a specific generative fuzzy system FuzzyS2S for sequence-to-sequence generative tasks. Part V evaluates the performance of the FuzzyS2S model in machine translation, summary generation, and code generation tasks. Part VI provides a conclusion and an outlook for future work.

## II. RELATED WORKS

### A. Generative Models

The advent of deep learning has led to significant advancements in generative modelling technology. In the field of NLP, sequence generation encompasses a range of sequence-to-sequence (seq2seq) [24] tasks such as machine translation, code generation, and summary generation. Early research primarily focus on Recurrent Neural Networks (RNNs) and long short-term memory (LSTM) networks [25]. However, the memory performance of these early seq2seq models degrades rapidly as sequence length increases, and they struggle to effectively differentiate tokens with varying importance [24].

To address these issues, Bahdanau et al. proposed the Attention Mechanism [26], which assigns different degrees of attention to tokens based on their importance by attentional scoring, thus alleviating the problem of long dependencies. Vaswani et al. proposed the Transformer model [4], which utilizes a multi-head attention mechanism and effectively parallelizes training. BERT [27] is a model based on the Transformer architecture that employs bidirectional encoding representation, and is used as the encoder in the autoencoder framework [28]. To support a variety of downstream tasks, Raffel et al. proposed the T5 [29], a Transformer-based pre-trained language model capable of performing tasks such as text classification, text generation, text summarization, and machine translation. ChatGPT is based on the GPT (Generative Pre-trained Transformer) family [5], including GPT-3.5 and GPT-4, which are capable of supporting a wide range of downstream tasks and enable interactive conversations.

Although current generative models have achieved remarkable performance, it is important to note that as their performance improves, the parameter scale of these models continues to grow exponentially. These parameters cannot be efficiently reused in other models [30], [31], [32], and their interpretability becomes increasingly challenging [15], [33]. Additionally, the learning processes of mainstream generative models are predominantly data-driven and lack knowledge-driven mechanisms [34],[35]. Therefore, it is highly beneficial to develop a novel generative framework to enhance the interpretability of generative models and establish a dual-driven mechanism by both data and knowledge.

### B. Fuzzy System

In 1965, American automatic control expert L.A. Zadeh proposed the concept of fuzzy sets, which led to the rapid development of fuzzy theory. The fuzzy system defines input, output and state variables on fuzzy sets, and serves as a representative uncertainty reasoning system. Fuzzy systems excel in solving nonlinear modelling problems and are now widely used in automatic control, pattern recognition, decision analysis, time series signal processing, and other tasks. Currently, there are two main branches of fuzzy systems: Mamdani fuzzy systems [36], [37] and TSK (Takagi-Sugeno-Kang) fuzzy systems [38], [39]. The TSK fuzzy system, In particular, has strong data-driven learning ability and good interpretability, which has garnered extensive attention and research in recent years.

The fuzzy reasoning in the fuzzy system is approximate and non-deterministic, with both premises and conclusions being inherently fuzzy. The fuzzy reasoning involves deriving reasoning conclusion from premises using hypothetical fuzzy propositions, known as fuzzy rules. Multiple fuzzy rules form the rule base. In the traditional and most widely used first-order TSK fuzzy system, the $k$-th rule of its fuzzy rule base, i.e., the hypothetical fuzzy proposition IF-THEN, is represented as shown in Eq. (1). In this representation, the IF part constitutes the fuzzy rule antecedent, and the THEN part constitutes the fuzzy rule consequent.

$$\text{IF: } x_1 \text{ is } A_1^k \wedge x_2 \text{ is } A_2^k \cdots \wedge x_D \text{ is } A_D^k$$
$$\text{THEN: } f^k(x) = p_0^k + p_1^k x_1 + \cdots + p_D^k x_D$$
$$k = 1,2,3,\cdots,K \quad (1)$$

In the above equation, $D$ is the dimension of the sample feature space, $K$ is the number of rules, $x_j (j = 1, 2, 3, ..., D)$ is the $j$th feature of the input vector $x$, and $A_j^k$ is the antecedent fuzzy set of the $k$-th fuzzy rule on the $j$-th feature of the input vector $x$. The symbol $\wedge$ denotes the fuzzy conjunction operator. The function $f^k(.)$ is the consequent processing function of the $k$-th rule, which, in this context, is a classical linear function used to derive the output of the $k$-th rule. The parameter $p_j^k (j = 1, 2, 3, ..., D)$ is the $j$-th parameter of the linear function adopted by the $k$-th rule consequent. Detailed construction of the antecedents and consequents of the TSK fuzzy system is given in Part 1 of the *Supplementary Materials*.

### III. GENERATIVE FUZZY SYSTEM

*A. Concept and Structure*

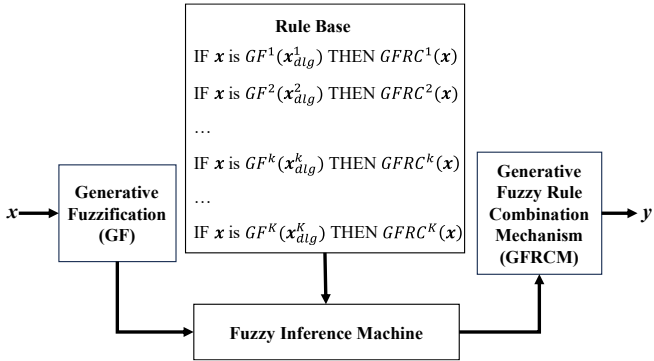

**Fig. 1.** Structure of Generative Fuzzy System Framework (GenFS)

**Definition 1: Generative Fuzzy System Framework (GenFS).** GenFS mainly consists of a generative fuzzification module, a generative fuzzy rule base, a fuzzy inference machine, and a generative fuzzy rule combination mechanism, as illustrated in Fig. 1. GenFS operates by processing input data through the generative fuzzification module and then reasoning with generative fuzzy rules to draw fuzzy inference conclusions under given preconditions. These inference conclusions are integrated through an efficient generative fuzzy rule combination mechanism, ultimately forming the system's output. GenFS is a novel framework that combines generative models and classical fuzzy systems, enabling fuzzy systems to handle complex generative tasks.

**Definition 2: Generative Fuzzification (GF).** It is the process of delegate election of fuzzy sets, aiming to transforming data into a more representative form and calculate the similarity between the inputs and the delegates.

**Definition 3: Generative Fuzzy Rule Consequent (GFRC).** GFRC is an intelligent fuzzy rule consequent that employs generative models as the consequent processing unit to enhance learning ability. GFRC can be represented as follows:
$$g^k = GFRC^k(x) \quad (2)$$
where $k = 1,2,3,...,K$, $K$ is the number of rules in the fuzzy system, $GFRC^k$ is the generative rule consequent processing unit of the $k$-th rule, and $g^k$ is the generative data of the $k$-th rule.

**Definition 4: Generative Fuzzy Rules.** Generative Fuzzy Rules are composed of antecedents with Generative Fuzzification and GFRCs. The $k$-th generative fuzzy rule of the rule base can be expressed as follows:
$$IF\ x\ is\ GF(x_{dlg}^k),THEN\ g^k = GFRC^k(x) \quad (3)$$
where $x$ is the input data and $GF(x_{dlg}^k)$ is the generative fuzzification function corresponding to the $k$-th rule antecedent, represented by $x_{dlg}^k$.

**Definition 5: Generative Fuzzy Rule Combination Mechanism (GFRCM).** GFRCM is an intelligent decision-making mechanism for generative tasks. Its purpose is to select applicable rule combination strategies to produce crisp and unambiguous results for different generative scenarios. When processing structured data with aligned features, GFRCM employs the weighted average method to combine the outputs of all the rules. For unstructured output data, GFRCM uses the maximum defuzzification method, selecting the output of the rule with the highest fire strength as the final result. Eq. (4) provides the generalized form of generative fuzzy rule combination mechanism:

$$y = GFRCM(G) = \begin{cases} \sum_{k=1}^{K} \tilde{\mu}^k * g^k, G \text{ is aligned} \\ g^{Argmax(\tilde{\mu})}, otherwise \end{cases} \quad (4)$$

where $G = \{g^1, g^2, g^3, ..., g^K\}$, $g^k$ is the output of $k$-th rule, $\tilde{\mu}^k$ is the fire strength of the $k$-th rule, and $\tilde{\mu} = [\tilde{\mu}^1, \tilde{\mu}^2, \tilde{\mu}^3, ..., \tilde{\mu}^K]$. When $G$ is structured and features are aligned, the fire strengths are used as weights to weight and sum all the rule outputs. Otherwise, the rule output with the maximum fire strength is selected as the final output, where $Argmax(\tilde{\mu})$ function is used to obtain the index value of the maximum fire strength.

*B. Construction of Generative Fuzzy Rule Antecedents*

The construction of generative fuzzy rule antecedents is a crucial task in modelling generative fuzzy systems. Classical fuzzy systems use the membership of the crisp data structured features corresponding to all the fuzzy sets as the fuzzification results. However, the inputs of generative tasks often vary in length after serialization, making classical fuzzification methods inadequate for directly handling unstructured data. Therefore, a novel generative fuzzification method for complex generative tasks.

The membership calculation in the generative fuzzification method requires solving two key problems: firstly, the election of delegates for fuzzy sets, and secondly, the calculation of similarity between the inputs and the delegates.

**Delegate Election.** A delegate of a set is a particular element of the set [40], [41]. The election of a delegate element simplifies the understanding and manipulation of the structure and properties of the set. In generative tasks, the delegate of a fuzzy set is a particular sample within the set that has a high degree of similarity in structure and properties with other samples. The method of delegate election can be based on expert experience or obtained through methods such as clustering. Each fuzzy set has its own individual delegate. The election of delegates can be expressed as Eq. (5).

$$X_{dlg} = DSM(X, K) \quad (5)$$

where $X_{dlg} = [x_{dlg}^1, x_{dlg}^2, \ldots, x_{dlg}^K]$ is a delegate set of $K$ rules in the system rule base, $X$ is the set of corresponding input data, $DSM$ denotes the method of the delegate election of the fuzzy set $X$.

**Similarity Calculation**. Similarity is usually expressed as a numerical value, where a larger value indicates greater similarity between two objects. In generative tasks, the inputs and delegates are usually unstructured data with unaligned features. Therefore, it is necessary to design flexible calculation methods tailored to different scenarios. Similarity can be calculated by data matching method, feature aligning method, and other specific methods. This can be presented in a generalized form, as shown in Eq. (6).

$$\mu^k = Sim(x, x_{dlg}^k) \quad (6)$$

where $x$ is the input data, $x_{dlg}^k$ is the delegate of the $k$-th fuzzy set, $k = 1,2,\ldots,K$, $K$ is the number of rules, $Sim(.)$ is a generalized method for calculating the similarity, $\mu^k$ is the similarity between $x$ and the delegate of the corresponding fuzzy set of the $k$-th rule, namely, the fire strength of $x$ about the $k$-th rule.

To stabilize numerical calculations and avoid feature bias, it is usually necessary to normalize similarity, as shown in Eq. (7).

$$\widetilde{\mu} = Norm(\mu) = \frac{\mu}{\sum_{k=1}^{K} \mu^k} \quad (7)$$

where $\widetilde{\mu} = [\tilde{\mu}^1, \tilde{\mu}^2, \tilde{\mu}^3, \ldots, \tilde{\mu}^K]$, $\mu = [\mu^1, \mu^2, \mu^3, \ldots, \mu^K]$.

### C. Learning and Optimization of GFRCs

The learning of GFRCs can be regarded as a typical machine learning task. In machine learning, the closer the predicted data distribution is to the real data distribution, the better the model's performance. In the generative task, calculating the difference between the target sequence and the generated sequence is key to the model's learning process. From a probability distribution perspective, the target sequence is considered the conditional probability distribution of the input sequence, and the difference between the generated sequence and the target sequence is derived from these two probability distributions. The cross-entropy loss function is widely used to measure this difference between probability distributions. From a sequence similarity perspective, the difference between sequences can be measured by their similarity, making the similarity loss function also applicable for sequence difference calculation. Depending on the scenarios, GenFS adopts different loss functions to calculate sequence differences. To prevent overfitting, an L2 regularization term is added to the loss function. The final loss function is generalized as follows:

$$\mathcal{L}_{GFRC}(\Theta) = diff(y, \hat{y}) + \gamma \|\Theta\|_2 \quad (8)$$

where $y$ is the ground true of target sequences in the generative task, $\hat{y}$ is the generative sequence, $diff(.)$ is a function to compute the difference between the target and generative sequences, $\|\Theta\|_2$ is the L2 regular term of the GFRC consequent parameters, and $\gamma (\gamma > 0)$ is a balancing parameter.

For generative tasks, sequence data is high-dimensional and sparse, with an uneven token frequency distribution where high-frequency and low-frequency tokens appear randomly. During training, reducing the learning rate for high-frequency tokens and increasing it for low-frequency tokens helps the model effectively learn information from both types of tokens. Therefore, GFRC needs to employ optimizers capable of adaptively adjusting the learning rate, such as AdaGrad, RMSprop, Adadelta and Adam [42], [43] to better approximate the optimal values of the model parameters and minimize the training loss. The generalized representation of the adaptive optimizer is shown in Eq. (9).

$$\Theta' = AdaOptim(\Theta, \ell_s) \quad (9)$$

where $AdaOptim(.)$ is the adaptive optimiser function, $\ell_s$ is the initialised value of the learning rate, and $\Theta'$ is the model parameter that has been computationally updated by the optimizer.

### D. Generalized Learning Algorithm for GenFS

In knowledge-based systems, a series of IF-THEN conditional statements can be used as rules to express knowledge. The antecedents' parameters can be derived from the prior knowledge of human experts or clustering techniques. The consequents acquire knowledge by learning the mapping relationship of sequences. The construction of generative fuzzy rules involves two main tasks: constructing the antecedents and learning the consequents' parameters.

The model parameters in the generative fuzzy system, denoted as $\Theta$, include the parameters of both the antecedents and consequents. The consequents can be initialized randomly or with pre-trained model parameters. Multiple consequents can be trained jointly. The generalized learning algorithm for GenFS is described in Algorithm I.

**Algorithm I:** Algorithm of GenFS

---

**Input**: Generative Input Sequence $X = x_1, x_2, \ldots, x_N$;
  Generative Target Sequence $Y = y_1, y_2, \ldots, y_M$;
  Number of rules: $K$;
  Batch size;
  Epoch: maximal number of epochs;
**Output**: GFRC parameters $\Theta$

---

**Procedure**:
// *Antecedent generation*
1: Use clustering or other technique to find the delegate of each fuzzy set in the rule antecedents, $X_{dlg} = x_{dlg}^1, x_{dlg}^2, \ldots, x_{dlg}^K$
// *Consequent generation*
2: **for** epoch=1 to Epoch **do**
3:   **for** each batch **do**
4:     Run Adam or other optimizers to minimize $\mathcal{L}_{GFRC}$ in Eq. (8)
5:     $\Theta \coloneqq \underset{\Theta}{\operatorname{argmin}} \mathcal{L}_{GFRC}$

6:    Calculate the training accuracy, validation accuracy, training loss, and validation loss for each batch.
7:  **end for**
8:  Calculate the average training accuracy, validation accuracy, training loss, and validation loss for each epoch according to the results obtained in Step 7 and save temporary optimal parameters Θ.
9:  **end for**
10: Output model parameters Θ with the optimal validation accuracy.

## IV. Sequence-to-Sequence Generation Based On GenFS Framework

In this section, we propose a specific generative fuzzy system based on the GenFS framework, tailored for sequence-to-sequence generation tasks. First, we define the specific problems involved. Then, we propose a novel multi-scale tokenizer, namely fuzzy tokenizer, for sequence splitting preprocessing. Furthermore, we propose an end-to-end generative model based on GenFS, named FuzzyS2S.

### A. Problem Definition

In NLP tasks such as machine translation, code generation, and summary generation, the text sequence is sliced into tokens by tokenizers, and these tokens follow the long-tailed distribution of Zipf's Law [44]. Zipf's Law is an empirical law that describes the relationship between token frequency and token ranking in natural language. The basic formulation of Zipf's Law is that in a large corpus, the token frequency is inversely proportional to its position in the token ranking list. Specifically, the law is expressed as:

$$f(r) \approx C/n \tag{10}$$

where $f(n)$ denotes the frequency of the token in the ranking list, $n$ denotes the rank of the token, and $C$ is a constant.

In natural language, the occurrence of tokens still conforms to Zipf's Law. While a small number of tokens occur with high frequency, the majority of tokens occur with low frequency. Thereby, this raises two problems that require resolution:

**Problem I**: A significant proportion of the tokens in sequences are low-frequency, which presents a challenge for the model learning the occurrence patterns of these tokens. Therefore, it is significant to reduce the number of low-frequency tokens and optimize the token distribution.

**Problem II**: There are significant variations in token distributions and sequence lengths across sequences, posing challenges for model fitting.

### B. Multiscale Tokenizer based on the priori expert knowledge: Fuzzy Tokenizer

To address the issue of low-frequency tokens (Problem I), we propose a specific multi-scale tokenizer, called fuzzy tokenizer. The fuzzy tokenizer is a fuzzy system based on multi-scale sub-word tokenizers [45] that enables the adaptive slicing of words at different scales. The architecture of the fuzzy tokenizer is illustrated in Fig. 2.

Given the original text sequence $s$, a preliminary token sequence $s'$ can be obtained after the basic tokenization:

$$s' = T_{basic}(s) = [x_1, x_2, x_3, \ldots, x_N] \tag{11}$$

where $T_{basic}$ is the basic tokenizer, which splits text sequences with spaces or punctuation marks as separators. $n$ is the total number of tokens of the sequence $s'$, $x_n$, $n = 1,2,3,\ldots,N$ is the $n$th token of the sequence $s'$.

The cosine distance algorithm is used in the antecedents of rules to calculate the similarity between the input tokens and the delegates of the fuzzy sets. This similarity is used to represent the fire strength of the fuzzy rules. Calculating the fire strength $\tilde{\mu}_n$ of token $x_n$ with respect to all rules can be expressed as follows:

$$\mu_n^k = Cosine(x_n, x_{dlg}^k) \tag{12}$$

$$\tilde{\boldsymbol{\mu}}_n = Norm(\boldsymbol{\mu}_n) \tag{13}$$

where $\mu_n^k$ denotes the fire strength of token $x_n$ with respect to the $k$th rule, i.e., $\mu_n = [\mu_n^1, \mu_n^2, \mu_n^3, \ldots, \mu_n^K]$, and the normalised fire strength $\tilde{\boldsymbol{\mu}}_n = [\tilde{\mu}_n^1, \tilde{\mu}_n^2, \tilde{\mu}_n^3, \ldots, \tilde{\mu}_n^K]$.

The consequents of the fuzzy tokenizer are sub-word tokenizers operating at different scales. The outputs are unstructured sub-word sequences. Fuzzy tokenizer selects the maximum defuzzification method to obtain the final result, as showed in Eq. (14).

$$\tilde{\boldsymbol{x}}_n = T_{fuzz}^{Argmax(\tilde{\boldsymbol{\mu}}_n)}(x_n) \tag{14}$$

where $\tilde{\boldsymbol{x}}_n = [x_{n1}, x_{n2}, x_{n3}, \ldots, x_{n\tau}]$ is the sub-word sequence after slicing of token $x_n$ and $\tau$ is the total number of sub-words. The fuzzy tokenizer splices the results of further slicing to obtain the final token sequence $\tilde{s}$.

$$\tilde{s} = concat(\tilde{\boldsymbol{x}}_1, \tilde{\boldsymbol{x}}_2, \tilde{\boldsymbol{x}}_3, \ldots, \tilde{\boldsymbol{x}}_N) \tag{15}$$

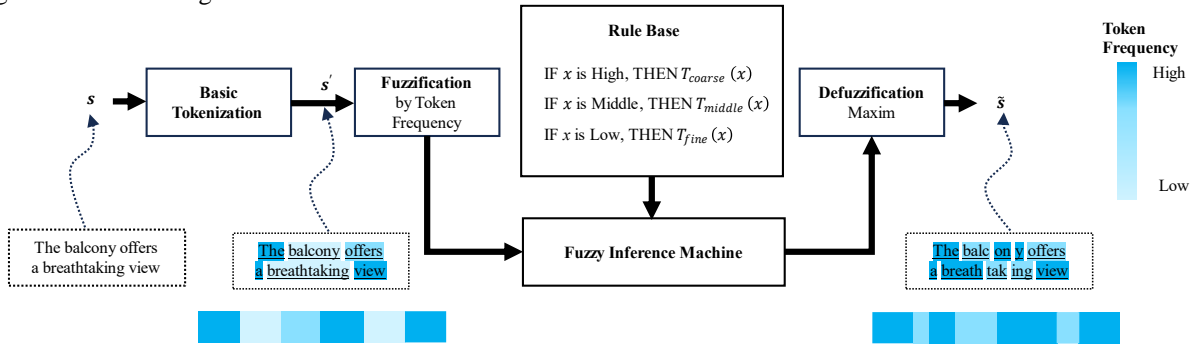

**Fig. 2.** Structure of Fuzzy Tokenizer

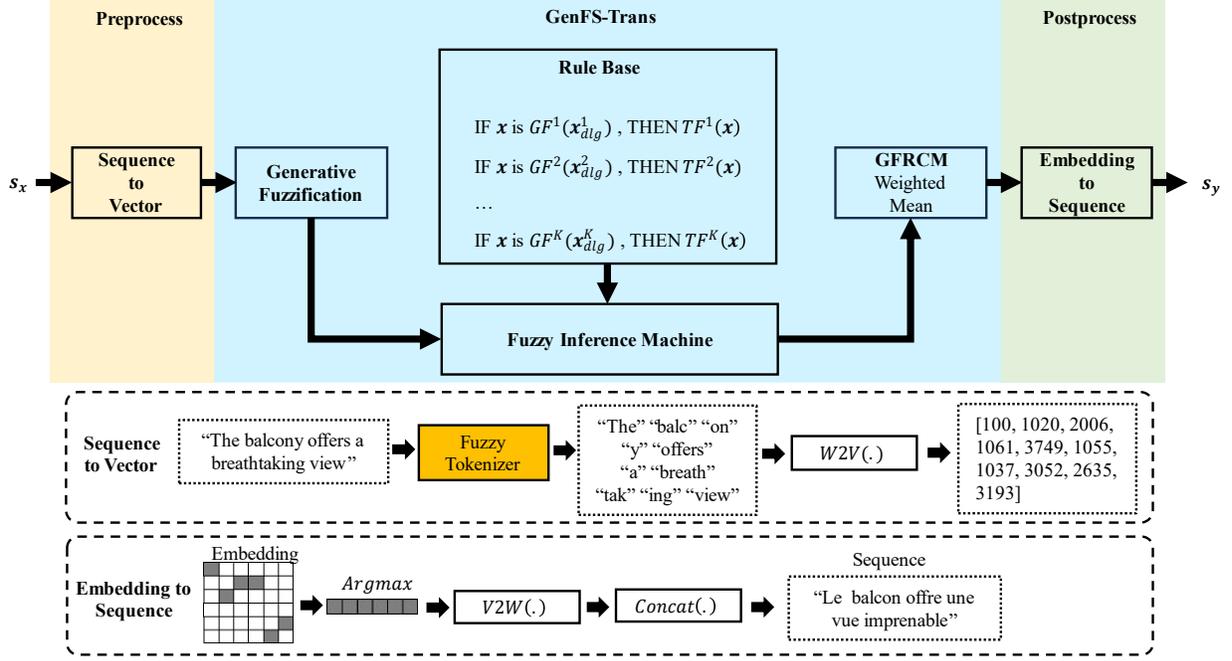

**Fig. 3.** Structure of FuzzyS2S, $TF^k$ is the Transformer processing unit of the $k$-th rule consequent, $s_x$ is the input sequence, and $s_y$ is the target sequence. the Preprocess module named *Sequence to Vector* is to implement the conversion from sequences to word vectors, and the Postprocess module named *Embedding to Sequence* is to convert the decoded word embeddings into the target sequences.

The fuzzy tokenizer has been designed to obtain more optimal splitting results based on fuzzy inference, considering the inherent fuzziness and uncertainty of the tokens at different scales. The fuzzy tokenizer combines the fuzzy system and the multi-scale learning. By integrating sub-word tokenizers of different scales, the fuzzy tokenizer leverages the universal approximator properties of the fuzzy system [46], [47], [48], [49] to approximate the token sequences with the best token frequency distribution and suitable sequence length.

*C. Sequence-to-sequence GenFS: FuzzyS2S*

This section presents an end-to-end sequence generative model based on GenFS, denoted as FuzzyS2S. In FuzzyS2S, we implement a special GenFS, using Transformer as the consequent processing units, denoted as GenFS-Trans. FuzzyS2S introduces two new modules: the Preprocess module, and the Postprocess module.

The three modules are described below:

**Preprocess Module**. This module transforms the original input sequences into vectors, reducing the low-frequency tokens in the sequences using the fuzzy tokenizer. In the preprocessing stages, the source sequence $s_x$ is passed through the fuzzy tokenizer to obtain the sequence of tokens $s'_x$ as shown in Eq. (16).

$$s'_x = T_{fuzzy}(s_x) = [x_1, x_2, x_3, \dots, x_\xi, \dots, x_N] \quad (16)$$

where $N$ is the number of tokens in the sequence, $x_\xi$ is the $\xi$th token in the sequence, $\xi = 1,2,3,\dots,N$. The sliced tokens are saved using the classical bag-of-words (BoW) model [50], [51], [52]. The input of BoW is the sequence of tokens, and the output is the word vector, which is an array of indexes of the storage locations, and the size of BoW is denoted as $N_{bag}$. $W2V$ represents the method of mapping the token sequences to vectors, while $V2W$ represents the method of mapping the vectors to the token sequences. There are mainly two special tokens in BoW, namely the start marker *bos* and the end marker *eos*, which are added at the start and end positions of the sequence, respectively. As shown in Eq. (17), here the sequence of tokens $s'_x$ is vectorised to obtain the word vector, which will further be passed as input to GenFS-Trans.

$$v_x = W2V(s'_x) \quad (17)$$

**GenFS-Trans Module**. This module is the core of FuzzyS2S. It begins by performing fuzzification on the input vectors, then conducts the fuzzy inference using the fuzzy inference machine, and finally fuses the results of all the rules via the GFRCM component. Specifically, as shown in Eqs. (18) and (19), the fuzzy set delegate is obtained through the unsupervised fuzzy clustering technology, and the similarity between the inputs and the delegates are computed using the cosine distance algorithm. The consequent processing units are Transformer. According to Eq. (21), the word vectors are transformed into word embeddings with the addition of positional encoding to preserve the positional information of the token in the sequence. Eq. (22) shows that the word embeddings output by Transformer are passed through a multilayer perceptron (MLP) to raise the feature dimensions to $N_{bag}$. Furthermore, in the GFRCM component of GenFS-Trans, the weighted average method is used to combine the results of all the rules. The fire strengths of the rules serve as weights, and the word embeddings are weighted and averaged to obtain the target word embedding $Em_y$ as shown in Eq. (23). Finally, the resulting embedding is compressed into the range of (0,1) to obtain the $H$ matrix through $Softmax$ function.

$$[s^1_{dlg}, s^2_{dlg}, \dots, s^k_{dlg}, \dots, s^K_{dlg}] = DSM_{FCM}(S_x, K) \quad (18)$$

$$\mu_s^k = Cosine(\boldsymbol{s}_x, \boldsymbol{s}_{dlg}^k) \quad (19)$$
$$\widetilde{\boldsymbol{\mu}}_s = Norm(\boldsymbol{\mu}_s) \quad (20)$$
$$\boldsymbol{Em}_x = Pos\left(Emb(B(\boldsymbol{v}_x))\right) \quad (21)$$
$$\boldsymbol{Em}_y = \sum_{k=1}^{K} \tilde{\mu}_s^k MLP(TF^k(\boldsymbol{Em}_x)) \quad (22)$$
$$\boldsymbol{H} = Softmax(\boldsymbol{Em}_y) \quad (23)$$

where $\boldsymbol{S}_x$ is the set of input sequences; $DSM_{FCM}(.)$ is a delegate election method based on Fuzzy C-Means[53]; the cosine distance algorithm $Cosine(.)$ is used to calculate the similarity between the input data and the delegates; $\boldsymbol{\mu}_s = [\mu_s^1, \mu_s^2, \mu_s^3, ..., \mu_s^K]$, normalized similarity $\widetilde{\boldsymbol{\mu}}_s = [\tilde{\mu}_s^1, \tilde{\mu}_s^2, \tilde{\mu}_s^3, ..., \tilde{\mu}_s^K]$; $v_x$ is the word vector of the source sequence, $B(.)$ is to add bos at the beginning of the word vector, $Emb(.)$ is to compute the word embedding of the word vector, $Pos(.)$ is to encode the position of the word embedding; $TF^k$ is the Transformer processing unit for the $k$th consequent, $\boldsymbol{Em}_x \in \mathbb{R}^{(N+1) \times Dim}$, $Dim$ is the dimensionality of the word embedding, $\boldsymbol{Em}_y \in R^{(M+1) \times N_{bag}}$, and $\boldsymbol{H}$ is the predictive probability matrix, $\boldsymbol{H} \in \mathbb{R}^{(M+1) \times N_{bag}}$.

**Postprocess Module**. This module is responsible for converting the word embeddings output from the GenFS-Trans module into the target sequence. According to Eq. (24), the process uses the $Argmax$ function to obtain the index of the maximum value of the probability for each dimension, and finally predicts $M + 1$ tokens.

$$s_{\hat{y}} = Concat\left(V2W(Argmax(\boldsymbol{H}))\right)$$
$$= Concat([\hat{y}_1, \hat{y}_2, \hat{y}_3, ..., \hat{y}_M, eos]) \quad (24)$$

where $V2W(.)$ is a vector-to-token conversion function and $Concat(.)$ is a function that splices an array of tokens into a sequence, FuzzyS2S stops the token prediction when the stop marker *eos* occurs.

## V. EXPERIMENT

### A. Datasets and Experiment Setting

The experiments follow the Train-Validation-Test (TVT) approach as outlined in [54]. The samples are divided into three parts: training sets, validation sets, and test sets.

The datasets include three categories: machine translation, summary generation, and code generation, totaling 12 datasets. The datasets for machine translation are WMT14 [54], Tatoeba [55], EUconst [56] and Ubuntu [56]. The datasets for summary generation are CNN/DM (CNN Daily Mail) [57], [58], SAMSum [59], XLSum[60] and BillSum [61]. The datasets for code generation are HS (HearthStone) [62], MTG (Magic the Game) [62], GEO (Geoquery) and Spider [63]. Detailed introductions of these datasets are provided in Part 2 of the *Supplementary Materials*.

The comparative methods in the experiments are Transformer [4], T5 [29] and CodeT5 [64]. Further details on these models can be found in Part 3(A) of the *Supplementary Materials*. The parameter settings for FuzzyS2S and the indicators for model evaluation, including ACC, BLEU [65], METEOR [66], ROUGE-1(R1), ROUGE-2(R2), and ROUGE-L(RL) [29] which are detailed in Part 3(B) and 3(C) of the *Supplementary Materials*.

### B. Analysis of Sequence Generation Experiments

**1) Analysis of Machine Translation Experiments**

The comparison of the machine translation performance metrics of FuzzyS2S and T5, as presented in Table I, reveals that FuzzyS2S outperforms T5 in terms of accuracy, with notable differences of 52.69 on the Tatoeba dataset and 52.64 on the EUconst dataset. Although FuzzyS2S's BLEU and METEOR metrics are lower than those of T5 on some datasets, it nevertheless shows a certain overall advantage. For the EUconst dataset, FuzzyS2S achieved a BLEU score of 1.7 higher than that of T5, while the METEOR metric indicated an 11.05-point advantage. The above results can be analyzed as follows:

(1) FuzzyS2S enhances accuracy by leveraging the interconnectivity between tokens of different scales. When the occurrence frequency of coarse-scale tokens is minimal, it becomes challenging for the model to capture their semantic information. By splitting the text sequence into fine-scale tokens, the occurrence frequency of these tokens increases.

(2) The generative consequents of FuzzyS2S are based on the classical Transformer, which employs absolute positional coding [4], [67]. In contrast, T5 employs relative positional coding [68]. Absolute positional coding encodes the positions of all input tokens. However, the relative position coding, as used by T5, considers the relative distance between the current token position and the position of the attended token when calculating the attention score [68]. This approach focuses more on the fluency of the sequences, which can lead to slightly poorer accuracy of the token prediction.

From Table I, FuzzyS2S exhibits better performance than the Transformer. This improvement can be attributed to the fusion of the GFRCs (Transformer) using the weighted averaged method, with fire strengths as weights. This method is similar to the naive soft attention mechanism [69], [70], [71], [72]. As the fire strength can adaptively change based on the inputs, the weighted fusion method functions as a dynamically adaptive soft-attention mechanism. This attention mechanism allows FuzzyS2S generally outperform the classical Transformer.

**2) Analysis of Summary Generation Experiments**

The results of the experimental analysis on the summary generation dataset are presented in Table II. This table demonstrates that FuzzyS2S outperforms Transformer on all datasets. Furthermore, it exhibits superior performance on three of the four datasets (SAMSum, XLSum and BillSum) compared to the T5 family, with an average of 5.23 higher on RL performance. These results can be analyzed in light of the following observations: Since the input for the summary generation task is an entire article or report, which constitutes a long sequence, there is a significant long-term dependency problem. One potential solution to this issue is the introduction of an attention mechanism. The generative rule consequent of FuzzyS2S is the Transformer with a multi-head attention mechanism [4]. Additionally, there is a layer of rule-level soft attention mechanism on top of all the generative rule consequents of FuzzyS2S. This demonstrates that FuzzyS2S is particularly well-suited for the summary generation task.

TABLE I
RESULTS OF MACHINE TRANSLATION EXPERIMENTS

| Models | WMT14 | Tatoeba | EUconst | Ubuntu |
|---|---|---|---|---|
| | ACC/BLEU/METEOR | ACC/BLEU/METEOR | ACC/BLEU/METEOR | ACC/BLEU/METEOR |
| **T5-Small** | 6.70/27.31/53.71 | 7.23/36.20/61.37 | 2.97/36.75/75.88 | 4.24/2.76/15.61 |
| **T5-Base** | 7.33/29.96/55.74 | 7.89/39.96/64.07 | 2.81/38.26/77.14 | 4.48/3.33/16.52 |
| **T5-Large** | 7.40/**30.92**/**56.8** | 8.01/**41.45**/65.19 | 2.87/38.59/77.42 | 5.65/3.90/17.43 |
| **Transformer** | 7.71/6.15/31.9 | 59.91/37.80/59.01 | 55.45/37.72/69.88 | 14.56/7.97/41.24 |
| **FuzzyS2S** | **8.49**/11.20/39.36 | **59.92**/37.83/**66.08** | **55.61/49.64/79.12** | **14.67/7.99/41.78** |

TABLE II
RESULTS OF THE SUMMARY GENERATION EXPERIMENTS

| Models | CNN/DM | SAMSum | XLSum | BillSum |
|---|---|---|---|---|
| | R1/R2/RL | R1/R2/RL | R1/R2/RL | R1/R2/RL |
| **T5-Small** | 41.12/19.56/38.35 | 23.48/6.10/18.69 | 17.84/4.96/12.97 | 24.89/10.23/17.76 |
| **T5-Base** | 42.05/20.34/39.40 | 24.14/6.85/18.79 | 18.16/5.08/12.65 | 25.45/11.76/18.77 |
| **T5-Large** | **42.50/20.68/39.94** | 27.36/9.09/22.09 | 23.14/7.14/16.04 | 28.64/13.67/21.75 |
| **Transformer** | 24.51/4.51/18.08 | 30.16/6.03/25.70 | 18.46/3.69/16.59 | 40.92/15.74/27.54 |
| **FuzzyS2S** | 32.10/12.14/23.24 | **35.13/13.26/28.25** | **23.76/7.28/16.88** | **42.83/20.26/30.45** |

TABLE III
RESULTS OF THE CODE GENERATION EXPERIMENTS

| Models | HS | MTG | GEO | Spider |
|---|---|---|---|---|
| | ACC/BLEU/METEOR | ACC/BLEU/METEOR | ACC/BLEU/METEOR | ACC/BLEU/METEOR |
| **CodeT5-Small** | 3.85/36.56/37.34 | 2.44/46.98/64.93 | 7.39/43.00/62.42 | 3.25/16.19/43.19 |
| **CodeT5-Base** | 3.55/63.56/72.14 | 2.64/60.01/71.57 | 7.88/66.21/77.57 | 5.39/26.55/54.08 |
| **CodeT5-Large** | 4.53/66.02/75.67 | 3.38/73.47/76.97 | 8.49/84.18/88.50 | 5.55/**26.98/57.49** |
| **Transformer** | 44.67/70.82/70.44 | 27.34/58.70/63.10 | 83.47/89.65/89.66 | 36.29/21.48/46.28 |
| **FuzzyS2S** | **48.89/72.14/76.15** | **30.30/73.98/76.98** | **87.33/91.66/92.54** | **36.35**/22.33/48.60 |

TABLE IV
RESULTS OF FUZZYS2S ABLATION TEST

| Model | EUconst | SAMSum | HS |
|---|---|---|---|
| | ACC/BLEU/METEOR | ACC/BLEU/METEOR | ACC/BLEU/METEOR |
| **FuzzyS2S** | **55.61/49.64/79.12** | **35.13/13.26/28.25** | **48.89/72.14/76.15** |
| **- Fuzzy Tokenizer** | 55.39/46.01/73.20 | 30.95/7.48/26.54 | 47.82/71.11/72.15 |
| **- GenFS-Trans** | 54.65/45.59/72.63 | 30.16/6.03/25.70 | 44.67/70.82/70.44 |

**3) Analysis of Summary Generation Experiments**

The results of the experimental analysis on the summary generation dataset are presented in Table II. This table demonstrates that FuzzyS2S outperforms Transformer on all datasets. Furthermore, it exhibits superior performance on three of the four datasets (SAMSum, XLSum and BillSum) compared to the T5 family, with an average of 5.23 higher on RL performance. These results can be analyzed in light of the following observations: Since the input for the summary generation task is an entire article or report, which constitutes a long sequence, there is a significant long-term dependency problem. One potential solution to this issue is the introduction of an attention mechanism. The generative rule consequent of FuzzyS2S is the Transformer with a multi-head attention mechanism [4]. Additionally, there is a layer of rule-level soft attention mechanism on top of all the generative rule consequents of FuzzyS2S. This demonstrates that FuzzyS2S is particularly well-suited for the summary generation task.

**4) Analysis of Code Generation Experiments**

The experimental results on the code generation dataset are shown in Table III. As shown in the table, the accuracy of FuzzyS2S is significantly better than that of CodeT5 on all datasets, with an average of 28.69. Compared with the classical Transformer, FuzzyS2S also demonstrates a higher accuracy, with an average of 2.77. On the HS, MTG, and GEO datasets, the fluency of FuzzyS2S is also better than or close to CodeT5.

CodeT5, developed based on T5 [64], utilizes relative positional encoding. Due to the strong correlations between tokens in natural language sequences, relative positional encoding is particularly effective, enhancing fluency when processing natural language. However, code sequences, unlike natural language, have weaker inter-word-element correlations. Thus, the CodeT5 method based on relative positional coding does not show a significant improvement in fluency compared to FuzzyS2S and Transformer.

*C. Ablation Test*

The ablation test of FuzzyS2S is carried out in this subsection, with partial results reported in Table IV. Comparing the results of the first row with the second row, it is evident that the fuzzy tokenizer contributes to the model performance enhancement on almost all the datasets. Furthermore, comparing the results of the second row with the third row, it is clear that GenFS-Trans enhances performance across all datasets. This enhancement is also evident in machine translation and summary generation tasks, validating the effectiveness of GenFS-Trans in contributing to FuzzyS2S's overall performance. For more analysis of ablation test, please refer to Part 4 of the *Supplementary Materials*.

*C. Convergence Analysis*

In particular, the convergence curves of FuzzyS2S when trained on EUconst, SAMSum, and HS datasets are illustrated in Fig. 4. From the figure, it can be observed that the loss of FuzzyS2S tends to converge after 20, 30 and 20 epochs of training on the aforementioned three datasets, respectively. The experimental results demonstrate that FuzzyS2S exhibits relatively good stability and convergence.

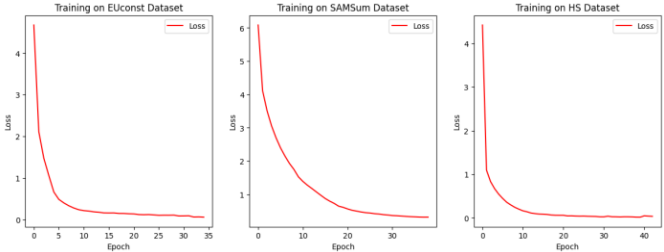

**Fig.4.** Convergence Analysis of FuzzyS2S on EUconst, SAMSum, HS Datasets

*D. Interpretability Analysis*

In this section, the interpretability of FuzzyS2S is analyzed and demonstrated through a case study. Fig. 5 provides an example of translating English sentences into French from the Tatoeba dataset. In the experiment, we configured the GenFS-Trans module in FuzzyS2S to contain three rules, each expressed as follows:

IF $s_x$ is *Long Sentence*, THEN $s_y = TF^{Long}(s_x)$;

IF $s_x$ is *Middle Sentence*, THEN $s_y = TF^{Middle}(s_x)$;

IF $s_x$ is *Short Sentence*, THEN $s_y = TF^{Low}(s_x)$;     (25)

The input English sentence in Fig. 5 has a length of 16, and its length and token frequency distribution characteristics are more similar to the description of the second rule. The actual similarity calculation indicates that the input sentence is more similar to the delegate in the *Middle Sentence* fuzzy set. Consequently, the result obtained through the GFRCM mechanism is consistent with the output of the *Middle Sentence* fuzzy rule. This example demonstrates that when the features of the input sentence align with the fuzzy term descriptions of the generative rule antecedent, the final output closely matches the output of the corresponding rule. This shows that the generative process of FuzzyS2S is semantically interpretable and also verifies the semantic interpretability of GenFS.

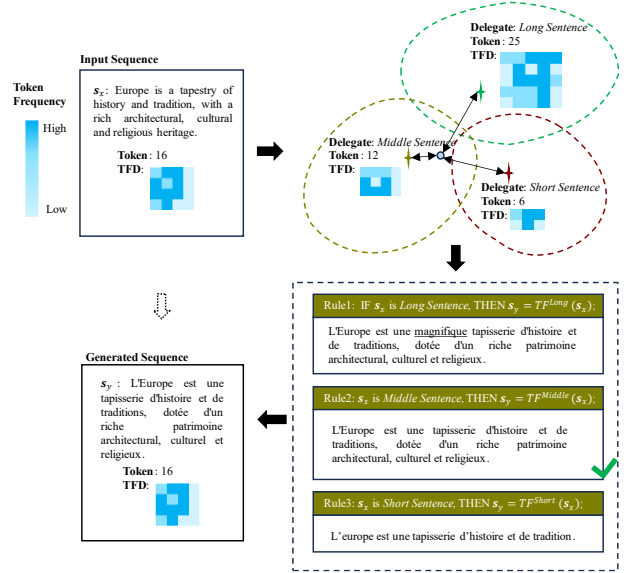

**Fig. 5.** An example of its translation of long, middle and short English sentences into French in FuzzyS2S, where the length of the delegate of *Long Sequence* fuzzy set is 25, the length of the delegate of *Middle Sequence* fuzzy set is 12, the length of the delegate of *Short Sequence* fuzzy set is 6, and Token Frequency Distribution denotes TFD.

*E. Other Analysis*

The fuzzy tokenizer is effectively capable of optimizing the token frequency distribution through multi-scale tokenization. Further analysis of the fuzzy tokenizer can be found in Part 5 of the *Supplementary Materials*. The number of rules is a key hyperparameter, and the performance of FuzzyS2S is subject to fluctuations based on the number of rules. An analysis of this hyperparameter is provided in Part 6 of the *Supplementary Materials*.

## VI. CONCLUSION

The learning process of current generative models is typically data-driven, lacking knowledge-driven mechanisms, making the modeling process a black-box system. To address these issues, we propose a novel generative modelling framework based on fuzzy systems, named GenFS. GenFS is capable of effectively handling complex generative modelling tasks. Furthermore, GenFS incorporates the classical fuzzy system's dual-driven learning mechanism of data and knowledge, offering semantical interpretability. For sequence-to-sequence generative learning tasks, we propose a novel end-to-end generative model based on GenFS framework, named FuzzyS2S, and verify its effectiveness in multiple scenarios, including machine translation, summary generation, and code generation. Our experimental results demonstrate that FuzzyS2S significantly outperforms the classical Transformer and surpasses the state-of-the-art T5 and CodeT5 models.

Although FuzzyS2S has demonstrated superior performance, there is still room for improvement. Firstly, GenFS is not yet capable of effectively supporting the modelling of multi-modal data, which limits its application scope. Secondly, while GenFS uses generative models as the processing units for the rule consequents, it retains some black box characteristics locally, indicating that interpretability needs to be further improved. In addition to the field of NLP, GenFS also has significant potential in the fields of image generation, audio generation, and video generation. In the future, further explorations will be conducted in order to address the aforementioned shortcomings.

REFERENCES

[1] B. Min *et al.*, "Recent advances in natural language processing via large pre-trained language models: A survey," *ACM Computing Surveys*, vol. 56, no. 2, pp. 1–40, 2023.

[2] W. X. Zhao *et al.*, "A survey of large language models," *arXiv preprint arXiv:2303.18223*, 2023.

[3] H. Touvron *et al.*, "Llama: Open and efficient foundation language models," *arXiv preprint arXiv:2302.13971*, 2023.

[4] A. Vaswani *et al.*, "Attention is all you need," *Advances in neural information processing systems*, vol. 30, 2017.

[5] T. Brown *et al.*, "Language models are few-shot learners," *Advances in neural information processing systems*, vol. 33, pp. 1877–1901, 2020.

[6] L. Ouyang *et al.*, "Training language models to follow instructions with human feedback," *Advances in Neural Information Processing Systems*, vol. 35, pp. 27730–27744, Dec. 2022.

[7] W. B. Knox and P. Stone, "Augmenting reinforcement learning with human feedback," in *ICML 2011 Workshop on New Developments in Imitation Learning*, 2011.

[8] D. Dai *et al.*, "Why can gpt learn in-context? language models implicitly perform gradient descent as meta-optimizers," *arXiv preprint arXiv:2212.10559*, 2022.

[9] J. Schulman, F. Wolski, P. Dhariwal, A. Radford, and O. Klimov, "Proximal policy optimization algorithms," *arXiv preprint arXiv:1707.06347*, 2017.

[10] A. Radford *et al.*, "Learning transferable visual models from natural language supervision," in *Proceedings Of The 38th International Conference On Machine Learning*, 2021, pp. 8748–8763.

[11] B. Li, K. Q. Weinberger, S. Belongie, V. Koltun, and R. Ranftl, "Language-driven semantic segmentation," *arXiv preprint arXiv:2201.03546*, 2022.

[12] Q. Huang, A. Jansen, J. Lee, R. Ganti, J. Y. Li, and D. P. Ellis, "Mulan: A joint embedding of music audio and natural language," *arXiv preprint arXiv:2208.12415*, 2022.

[13] A. Agostinelli *et al.*, "Musiclm: Generating music from text," *arXiv preprint arXiv:2301.11325*, 2023.

[14] Z. Borsos *et al.*, "Audiolm: a language modeling approach to audio generation," *IEEE/ACM Transactions on Audio, Speech, and Language Processing*, 2023.

[15] C. Rudin, "Stop explaining black box machine learning models for high stakes decisions and use interpretable models instead," *Nature Machine Intelligence*, vol. 1, no. 5, pp. 206–215, May 2019, doi: 10.1038/s42256-019-0048-x.

[16] S. Guillaume, "Designing fuzzy inference systems from data: An interpretability-oriented review," *IEEE Trans. Fuzzy Syst.*, vol. 9, no. 3, pp. 426–443, Jun. 2001, doi: 10.1109/91.928739.

[17] Y. Zhang, H. Ishibuchi, and S. Wang, "Deep Takagi–Sugeno–Kang Fuzzy Classifier With Shared Linguistic Fuzzy Rules," *IEEE Transactions on Fuzzy Systems*, vol. 26, no. 3, pp. 1535–1549, Jun. 2018, doi: 10.1109/TFUZZ.2017.2729507.

[18] T. Zhou, F.-L. Chung, and S. Wang, "Deep TSK Fuzzy Classifier With Stacked Generalization and Triply Concise Interpretability Guarantee for Large Data," *IEEE Transactions on Fuzzy Systems*, vol. 25, no. 5, pp. 1207–1221, Oct. 2017, doi: 10.1109/TFUZZ.2016.2604003.

[19] V. Singh, R. Dev, N. K. Dhar, P. Agrawal, and N. K. Verma, "Adaptive Type-2 Fuzzy Approach for Filtering Salt and Pepper Noise in Grayscale Images," *IEEE Trans. Fuzzy Syst.*, vol. 26, no. 5, pp. 3170–3176, Oct. 2018, doi: 10.1109/TFUZZ.2018.2805289.

[20] X. Ma *et al.*, "Deep Image Feature Learning With Fuzzy Rules," *IEEE Trans. Emerg. Top. Comput. Intell.*, pp. 1–14, 2023, doi: 10.1109/TETCI.2023.3259447.

[21] X. Tian *et al.*, "Deep Multi-View Feature Learning for EEG-Based Epileptic Seizure Detection," *IEEE Transactions on Neural Systems and Rehabilitation Engineering*, vol. 27, no. 10, pp. 1962–1972, Oct. 2019, doi: 10.1109/TNSRE.2019.2940485.

[22] Z. Deng, Y. Jiang, H. Ishibuchi, K.-S. Choi, and S. Wang, "Enhanced Knowledge-Leverage-Based TSK Fuzzy System Modeling for Inductive Transfer Learning," *ACM Trans. Intell. Syst. Technol.*, vol. 8, no. 1, pp. 1–21, Jan. 2017, doi: 10.1145/2903725.

[23] P. Xu, Z. Deng, J. Wang, Q. Zhang, K.-S. Choi, and S. Wang, "Transfer Representation Learning With TSK Fuzzy System," *IEEE Transactions on Fuzzy Systems*, vol. 29, no. 3, pp. 649–663, Mar. 2021, doi: 10.1109/TFUZZ.2019.2958299.

[24] I. Sutskever, O. Vinyals, and Q. V. Le, "Sequence to Sequence Learning with Neural Networks," in *Advances in Neural Information Processing Systems*, Curran Associates, Inc., 2014. Accessed: Feb. 10, 2023. [Online]. Available: https://proceedings.neurips.cc/paper/2014/hash/a14ac55a4f27472c5d894ec1c3c743d2-Abstract.html

[25] S. Hochreiter and J. Schmidhuber, "Long Short-Term Memory," *Neural Computation*, vol. 9, no. 8, pp. 1735–1780, Nov. 1997, doi: 10.1162/neco.1997.9.8.1735.

[26] J. K. Chorowski, D. Bahdanau, D. Serdyuk, K. Cho, and Y. Bengio, "Attention-based models for speech recognition," *Advances in neural information processing systems*, vol. 28, 2015.

[27] J. Devlin, M.-W. Chang, K. Lee, and K. Toutanova, "BERT: Pre-training of Deep Bidirectional Transformers for Language Understanding," in *Proceedings of the 2019 Conference of the North American Chapter of the Association for Computational Linguistics: Human Language Technologies, Volume 1 (Long and Short*

[28] D. Bank, N. Koenigstein, and R. Giryes, "Autoencoders," Apr. 03, 2021, *arXiv*: arXiv:2003.05991. Accessed: Jul. 04, 2023. [Online]. Available: http://arxiv.org/abs/2003.05991

[29] C. Raffel *et al.*, "Exploring the Limits of Transfer Learning with a Unified Text-to-Text Transformer," *Journal of Machine Learning Research*, vol. 21, no. 140, pp. 1–67, 2020, Accessed: May 13, 2023. [Online]. Available: http://jmlr.org/papers/v21/20-074.html

[30] C. Tan, F. Sun, T. Kong, W. Zhang, C. Yang, and C. Liu, "A Survey on Deep Transfer Learning," in *Artificial Neural Networks and Machine Learning – ICANN 2018*, vol. 11141, 2018, pp. 270–279.

[31] Y. Bengio, "Deep learning of representations for unsupervised and transfer learning," in *Proceedings of ICML workshop on unsupervised and transfer learning*, 2012, pp. 17–36.

[32] G. Mesnil *et al.*, "Unsupervised and transfer learning challenge: a deep learning approach," in *Proceedings of ICML Workshop on Unsupervised and Transfer Learning*, 2012, pp. 97–110.

[33] M. Du, N. Liu, and X. Hu, "Techniques for interpretable machine learning," *Communications of the ACM*, vol. 63, no. 1, pp. 68–77, Dec. 2019, doi: 10.1145/3359786.

[34] K. Chen, M. Yang, T. Zhao, and M. Zhang, "Data-Driven Fuzzy Target-Side Representation for Intelligent Translation System," *IEEE Transactions on Fuzzy Systems*, vol. 30, no. 11, pp. 4568–4577, Nov. 2022, doi: 10.1109/TFUZZ.2022.3167129.

[35] A. Porwal, E. J. M. Carranza, and M. Hale, "Knowledge-Driven and Data-Driven Fuzzy Models for Predictive Mineral Potential Mapping," *Natural Resources Research*, 2003.

[36] Mamdani, "Application of Fuzzy Logic to Approximate Reasoning Using Linguistic Synthesis," *IEEE Transactions on Computers*, vol. C–26, no. 12, pp. 1182–1191, Dec. 1977, doi: 10.1109/TC.1977.1674779.

[37] P. M. Larsen, "Industrial applications of fuzzy logic control," *International Journal of Man-Machine Studies*, vol. 12, no. 1, pp. 3–10, 1980.

[38] T. Takagi and M. Sugeno, "Fuzzy identification of systems and its applications to modeling and control," *IEEE Transactions on Systems, Man, and Cybernetics*, vol. SMC-15, no. 1, pp. 116–132, Jan. 1985, doi: 10.1109/TSMC.1985.6313399.

[39] M. Sugeno and G. T. Kang, "Structure identification of fuzzy model," *Fuzzy Sets and Systems*, vol. 28, no. 1, pp. 15–33, Oct. 1988, doi: 10.1016/0165-0114(88)90113-3.

[40] P. Hall, "On Representatives of Subsets," in *Classic Papers in Combinatorics*, 1987, pp. 58–62. doi: 10.1007/978-0-8176-4842-8_4.

[41] M. Hall Jr, "Distinct representatives of subsets," *Bulletin of the American Mathematical Society*, vol. 54, no. 10, pp. 922–926, 1948.

[42] D. P. Kingma and J. Ba, "Adam: A method for stochastic optimization," *arXiv preprint arXiv:1412.6980*, 2014.

[43] I. K. M. Jais, A. R. Ismail, and S. Q. Nisa, "Adam Optimization Algorithm for Wide and Deep Neural Network," *Kno. Eng. Da. Sc.*, vol. 2, no. 1, p. 41, Jun. 2019, doi: 10.17977/um018v2i12019p41-46.

[44] D. M. W. Powers, "Applications and explanations of Zipf's law," in *Proceedings of the Joint Conferences on New Methods in Language Processing and Computational Natural Language Learning - NeMLaP3/CoNLL '98*, Sydney, Australia: Association for Computational Linguistics, 1998, p. 151. doi: 10.3115/1603899.1603924.

[45] R. Sennrich, B. Haddow, and A. Birch, "Neural Machine Translation of Rare Words with Subword Units," Jun. 10, 2016, *arXiv*: arXiv:1508.07909. Accessed: Jun. 05, 2023. [Online]. Available: http://arxiv.org/abs/1508.07909

[46] Hao Ying, "General SISO Takagi-Sugeno fuzzy systems with linear rule consequent are universal approximators," *IEEE Trans. Fuzzy Syst.*, vol. 6, no. 4, pp. 582–587, Nov. 1998, doi: 10.1109/91.728456.

[47] X.-J. Zeng and M. G. Singh, "Approximation accuracy analysis of fuzzy systems as function approximators," *IEEE Transactions on fuzzy systems*, vol. 4, no. 1, pp. 44–63, 1996.

[48] B. Kosko, "Fuzzy systems as universal approximators," *IEEE transactions on computers*, vol. 43, no. 11, pp. 1329–1333, 1994.

[49] L.-X. Wang, "Fuzzy systems are universal approximators," in *[1992 proceedings] IEEE international conference on fuzzy systems*, 1992, pp. 1163–1170.

[50] W. A. Qader, M. M. Ameen, and B. I. Ahmed, "An Overview of Bag of Words;Importance, Implementation, Applications, and Challenges," in *2019 International Engineering Conference (IEC)*, Erbil, Iraq: IEEE, Jun. 2019, pp. 200–204. doi: 10.1109/IEC47844.2019.8950616.

[51] C.-F. Tsai, "Bag-of-Words Representation in Image Annotation: A Review," *ISRN Artificial Intelligence*, vol. 2012, pp. 1–19, Nov. 2012, doi: 10.5402/2012/376804.

[52] Y. Zhang, R. Jin, and Z.-H. Zhou, "Understanding bag-of-words model: a statistical framework," *Int. J. Mach. Learn. & Cyber.*, vol. 1, no. 1–4, pp. 43–52, Dec. 2010, doi: 10.1007/s13042-010-0001-0.

[53] J. C. Bezdek, R. Ehrlich, and W. Full, "FCM: The fuzzy c-means clustering algorithm," *Computers & Geosciences*, vol. 10, no. 2–3, pp. 191–203, Jan. 1984, doi: 10.1016/0098-3004(84)90020-7.

[54] O. Bojar *et al.*, "Findings of the 2014 Workshop on Statistical Machine Translation," in *Proceedings of the Ninth Workshop on Statistical Machine Translation*, Baltimore, Maryland, USA: Association for Computational Linguistics, Jun. 2014, pp. 12–58. doi: 10.3115/v1/W14-3302.

[55] J. Tiedemann, "The Tatoeba Translation Challenge – Realistic Data Sets for Low Resource and Multilingual MT," in *Proceedings of the Fifth Conference on Machine Translation*, Online: Association for Computational Linguistics, Nov. 2020, pp. 1174–1182. Accessed: May 15, 2023. [Online]. Available: https://aclanthology.org/2020.wmt-1.139

[56] J. Tiedemann, "Parallel Data, Tools and Interfaces in OPUS," in *Proceedings of the Eighth International Conference on Language Resources and Evaluation (LREC'12)*, Istanbul, Turkey: European Language Resources Association (ELRA), May 2012, pp. 2214–2218. Accessed: Jun. 28, 2023. [Online]. Available: http://www.lrec-conf.org/proceedings/lrec2012/pdf/463_Paper.pdf

[57] A. See, P. J. Liu, and C. D. Manning, "Get to the point: Summarization with pointer-generator networks," *arXiv preprint arXiv:1704.04368*, 2017.

[58] K. M. Hermann *et al.*, "Teaching machines to read and comprehend," *Advances in neural information processing systems*, vol. 28, 2015.

[59] B. Gliwa, I. Mochol, M. Biesek, and A. Wawer, "SAMSum Corpus: A Human-annotated Dialogue Dataset for Abstractive Summarization," in *Proceedings of the 2nd Workshop on New Frontiers in Summarization*, Hong Kong, China: Association for Computational Linguistics, Nov. 2019, pp. 70–79. doi: 10.18653/v1/D19-5409.

[60] T. Hasan *et al.*, "XL-Sum: Large-Scale Multilingual Abstractive Summarization for 44 Languages," in *Findings of the Association for Computational Linguistics: ACL-IJCNLP 2021*, Online: Association for Computational Linguistics, Aug. 2021, pp. 4693–4703. doi: 10.18653/v1/2021.findings-acl.413.

[61] A. Kornilova and V. Eidelman, "BillSum: A Corpus for Automatic Summarization of US Legislation," in *Proceedings of the 2nd Workshop on New Frontiers in Summarization*, Hong Kong, China: Association for Computational Linguistics, Nov. 2019, pp. 48–56. doi: 10.18653/v1/D19-5406.

[62] W. Ling *et al.*, "Latent predictor networks for code generation," *arXiv preprint arXiv:1603.06744*, 2016.

[63] T. Yu *et al.*, "Spider: A large-scale human-labeled dataset for complex and cross-domain semantic parsing and text-to-sql task," *arXiv preprint arXiv:1809.08887*, 2018.

[64] Y. Wang, W. Wang, S. Joty, and S. C. Hoi, "Codet5: Identifier-aware unified pre-trained encoder-decoder models for code understanding and generation," *arXiv preprint arXiv:2109.00859*, 2021.

[65] K. Papineni, S. Roukos, T. Ward, and W.-J. Zhu, "Bleu: a Method for Automatic Evaluation of Machine Translation," in *Proceedings of the 40th Annual Meeting of the Association for Computational Linguistics*, Philadelphia, Pennsylvania, USA: Association for Computational Linguistics, Jul. 2002, pp. 311–318. doi: 10.3115/1073083.1073135.

[66] A. Lavie and A. Agarwal, "Meteor: an automatic metric for MT evaluation with high levels of correlation with human judgments," in *Proceedings of the Second Workshop on Statistical Machine Translation - StatMT '07*, Prague, Czech Republic: Association for Computational Linguistics, 2007, pp. 228–231. doi: 10.3115/1626355.1626389.

[67] P. Dufter, M. Schmitt, and H. Schütze, "Position Information in Transformers: An Overview," *Computational Linguistics*, vol. 48, no. 3, pp. 733–763, Sep. 2022, doi: 10.1162/coli_a_00445.

[68] P. Shaw, J. Uszkoreit, and A. Vaswani, "Self-attention with relative position representations," *arXiv preprint arXiv:1803.02155*, 2018.

[69] Z. Niu, G. Zhong, and H. Yu, "A review on the attention mechanism of deep learning," *Neurocomputing*, vol. 452, pp. 48–62, Sep. 2021, doi: 10.1016/j.neucom.2021.03.091.

[70] S. Chaudhari, V. Mithal, G. Polatkan, and R. Ramanath, "An Attentive Survey of Attention Models," *ACM Trans. Intell. Syst. Technol.*, vol. 12, no. 5, pp. 1–32, Oct. 2021, doi: 10.1145/3465055.

[71] K. Xu *et al.*, "Show, attend and tell: neural image caption generation with visual attention," in *Proceedings of the 32nd International Conference on International Conference on Machine Learning - Volume 37*, in ICML'15. Lille, France: JMLR.org, Jul. 2015, pp. 2048–2057.

[72] A. Galassi, M. Lippi, and P. Torroni, "Attention in natural language processing," *IEEE transactions on neural networks and learning systems*, vol. 32, no. 10, pp. 4291–4308, 2020.